\documentclass[runningheads]{llncs}

\usepackage[mobile]{eccv}

\usepackage{eccvabbrv}
\usepackage{stmaryrd}
\usepackage{appendix}
\usepackage{multirow}
\usepackage{graphicx}
\usepackage{booktabs}
\usepackage{amsmath}
\usepackage{pifont}
\usepackage[accsupp]{axessibility}  

\usepackage[pagebackref,breaklinks,colorlinks,citecolor=eccvblue]{hyperref}

\usepackage{orcidlink}

\begin{document}

\title{ST-LDM: A Universal Framework for Text-Grounded Object Generation in Real Images} 


\author{Xiangtian Xue\inst{1,3}\and
	Jiasong Wu\inst{1,3}\thanks{Corresponding author}\and
	Youyong Kong\inst{1,3} \and
	Lotfi Senhadji\inst{2,3} \and
	Huazhong Shu\inst{1,3} }

\authorrunning{X. Xue et al.}

\institute{Key Laboratory of New Generation Artificial Intelligence Technology and Its Interdisciplinary Applications, Southeast University \and
	Laboratoire Traitement du Signal et de l'Image, Univ Rennes \and
	Centre de Recherche en Information Biom\`{e}dicale Sino-fran\c{c}ais (CRIBs)\\
	\email{xxt@seu.edu.cn}, \email{jswu@seu.edu.cn}}

\maketitle

\begin{abstract}
 We present a novel image editing scenario termed Text-grounded Object Generation (TOG), defined as generating a new object in the real image spatially conditioned by textual descriptions. Existing diffusion models exhibit limitations of spatial perception in complex real-world scenes, relying on additional modalities to enforce constraints, and TOG imposes heightened challenges on scene comprehension under the weak supervision of linguistic information. We propose a universal framework ST-LDM based on Swin-Transformer, which can be integrated into any latent diffusion model with training-free backward guidance. ST-LDM encompasses a global-perceptual autoencoder with adaptable compression scales and hierarchical visual features, parallel with deformable multimodal transformer to generate region-wise guidance for the subsequent denoising process. We transcend the limitation of traditional attention mechanisms that only focus on existing visual features by introducing deformable feature alignment to hierarchically refine spatial positioning fused with multi-scale visual and linguistic information. Extensive Experiments demonstrate that our model enhances the localization of attention mechanisms while preserving the generative capabilities inherent to diffusion models.
\end{abstract}

\section{Introduction}
\label{sec:intro}

Controllable text2img editing has demonstrated remarkable performance with ongoing exploration of diffusion models~\cite{10203146,NEURIPS2022_a0a53fef,Ruiz_2023_CVPR}, and recent studies have been predominantly directed towards enhancing the controllability of language in editing the appearance of objects~\cite{Kawar_2023_CVPR,hertz2022prompttoprompt}. As for generating new objects in real images, additional modalities such as bounding boxes and keypoints have to be introduced to explicitly regulate their spatial placement~\cite{li2023gligen}. 

The aforementioned modalities provide precise spatial constraints, but it is noteworthy that language-based instruction serves as the most fundamental and convenient interactive modality in real-world scenarios (\eg chatting with a robot commonly using language as the sole mode of interaction, conducting similar manipulations on extensive image datasets and dynamic video sequences). Moreover, language affords a considerable degree of creative flexibility while adhering to the principles of spatial logic. This enhances the naturalness of edited images, as the model can estimate the most suitable generation location by comprehending multi-modal information.

\begin{figure}[tb]
	\centering
	\includegraphics[width=1\textwidth]{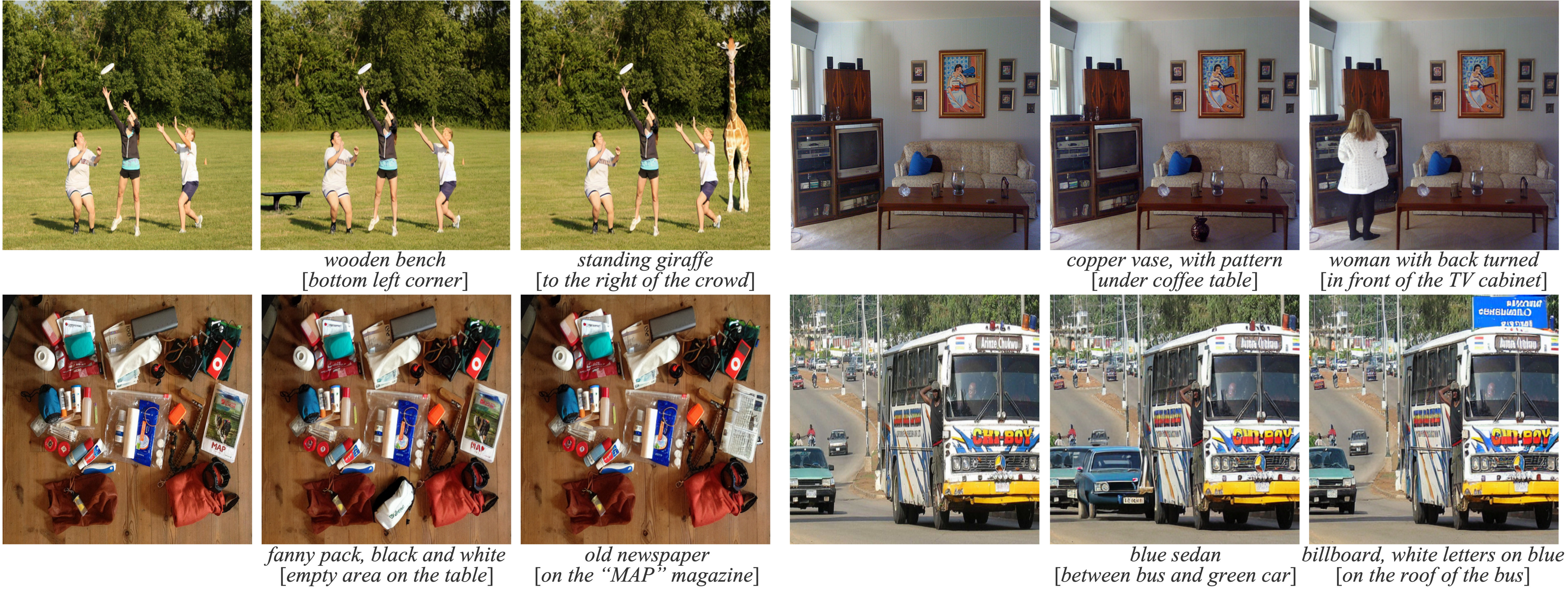}
	\caption{Text-grounded object generation in real images. Our method can generate the rational object within real-world input images based on textual descriptions encompassing its visual characteristic and spatial attribute. The generated object seamlessly integrates with the original image, exhibiting a visually harmonious presence while preserving the contextual semantics of the original scene (\eg consistency in shadow (top left),  geometry (top right), illumination (bottom left), occlusion (bottom right)).}
		\label{fig:onecol3}
		\vspace{-0.5cm}
\end{figure}

We propose a new image editing scenario named Text-grounded Object Generation (TOG), which aims to generate objects in real scenarios according to natural language expressions (\cref{fig:onecol3}). Concretely, given an image in conjunction with textual descriptions detailing visual and spatial attributes of the generated object, the task entails generating a rational object in complex scenarios. TOG embodies the dual characteristics of text2img generation and editing tasks. The generated object must adhere to textual descriptions, while simultaneously ensuring that the edited image maintains high fidelity with irrelevant areas remaining consistent with original semantics.

Although diffusion models have demonstrated notable creative capabilities in image generation, they fall short in spatial perception and orientation, particularly when dealing with intricate real-world scenes~\cite{Yang2022DiffusionMA}. The prevalent approaches to address this limitation involve introducing additional conditions during the denoising process to control the placement of the generated object. However, a fundamental paradox arises in our specific scenario: how to locate an object which is absent from the original image. This emphasizes the primary challenge, that is determining the optimal area for object generation under the weak supervision of texts. The objective is to ensure that the generated object aligns semantically with the contextual information contained within the image.

In this paper, we present ST-LDM, a universal framework with hierarchical Swin-Transformer (ST) blocks~\cite{Liu_2021_ICCV}, amenable for integration into latent diffusion models (LDM) via training-free backward guidance. ST-LDM comprises two parallel branches with similar structures, dedicated respectively to adaptable latent representation and spatial attention guidance. For latent representation, the Transformer-based autoencoder facilitates multi-resolution compression through a low-cost trainable module while preserving the formidable modeling capacity inherent to the pre-trained Swin-T. For spatial guidance, we propose deformable feature alignment in multimodal fusion Transformer to enhance the refinement of grounded regions by incorporating information from multi-scale visual and linguistic features. ST-LDM provides visually coherent location guidance in pixel space which can be integrated to the denoising U-Net of LDMs with region-wise backpropagation. 

To quantify the effectiveness of our proposed approach, we compare it with text-prompted image editing models with diverse conditions based on our constructed benchmark. Experiments demonstrate that our model enhances the localization capability of Swin-Transformer while preserving the generative proficiency inherent in diffusion models. The main contributions of our work are highlighted as follows: 
\begin{itemize}
	\item  We present a novel image editing scenario termed Text-grounded Object Generation (TOG), which is defined as generating objects in real images grounded by textual descriptions. 
	\item  We further propose a universal framework ST-LDM with deformable feature alignment to hierarchically refine spatial guidance, which can be integrated into diffusion models with training-free region-wise backpropagation.
	\item  Quantitive comparisons with related works on our constructed benchmark and qualitative visualizations substantiate the robust editing and generative capabilities according to textual conditions.
\end{itemize}

\section{Related Work}
\label{sec:related}
\subsection{Spatially Controllable Object Generation}
Recent studies have employed mandatory modalities to exert control over the spatial distribution of generated objects~\cite{fan2022frido,yang2023reco,chen2023trainingfree}. Make-A-Scene~\cite{gafni2022makeascene} introduces implicit conditioning over optionally controlled scene tokens derived from segmentation maps with fixed labels to control the layout of the image. SpaText~\cite{Avrahami2023} further proposes a spatial-textual representation that encapsulates the semantic attributes and structural characteristics of each segment. GLIGEN~\cite{li2023gligen} extends grounding conditions to bounding boxes, reference images, keypoints, \etc, which are injected into new trainable layers of the denoising process via gated self-attention. Nevertheless, these modalities are cumbersome when applied to large-scale operations involving numerous images and videos. The utilization of text-based spatial position descriptions serves to mitigate rigid constraints, thereby affording models a heightened capacity for comprehending visual information.

\subsection{Diffusion-Based Semantic Image Editing}
Image editing emphasizes modifying the visual entity consistent with textual descriptions while preserving the entity-irrelevant regions. Diffusion models can be easily adapted for inpainting when given a mask input since they iteratively refine an image starting from random noise~\cite{nichol2022glide,9880056}. Recent approaches endeavor to obviate the necessity of additional inputs such as image masks or additional views of the object~\cite{hertz2023prompttoprompt,9880101,chen2023trainingfree}. DiffEdit~\cite{couairon2023diffedit} automatically generates a mask accentuating regions of the input image necessitating modification, by contrasting predictions of a diffusion model conditioned on different text prompts. Imagic~\cite{Kawar_2023_CVPR} produces a text embedding that aligns with both the input image and the target text while fine-tuning the diffusion model to capture the image-specific appearance. These attempts are experimented on images with obvious foreground objects and simple semantics. However, when applied to complex scenes characterized by a multiplicity of objects and spatial relationships, their efficacy in localizing objects exhibits a discernible diminishment.

\subsection{Vision-and-Language Transformer}
The initial phase of our proposed method involves locating visual regions of interest guided by language, and Transformer~\cite{NIPS2017_3f5ee243} has demonstrated noteworthy performance at the interaction of vision and language. Ding \etal ~\cite{Ding_2021_ICCV} introduce Transformer and multi-head attention to construct an encoder-decoder attention mechanism architecture. LAVT~\cite{Yang_2022_CVPR} achieves better cross-modal alignments through the early fusion of linguistic and visual features in a vision Transformer encoder network. SLViT~\cite{OuYang2023SLViTSL} employs the variance of cross-modal correlations between adjacent stages to identify regions of high uncertainty and refine features of these regions with complementing information from multiple stages. While Transformer facilitates the integration of linguistic and visual modalities through cross-attention mechanisms, it encounters challenges stemming from the absence of target objects within the image, which is a distinctive characteristic of the TOG task.

\section{Method}
\begin{figure}[tb]
	\centering
	\includegraphics[width=1\linewidth]{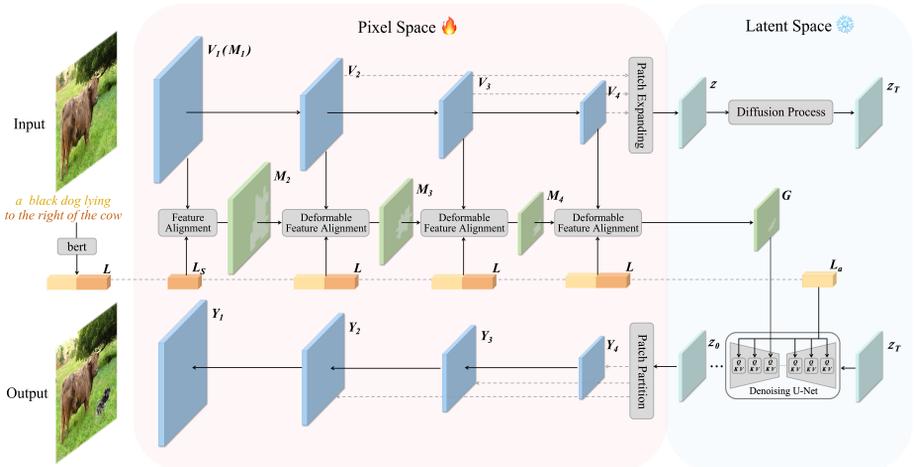}
	\caption{The basic framework of the proposed ST-LDM. Given a real image and text descriptions, we initially encode them into visual features $V_i$ and text embedding $L$, where $L$  comprises appearance prompt $L_a$ and spatial conditions $L_s$. We employ the Swin-T structure to hierarchically integrate $L$ into $V_i$, where the multimodal map $M_i$ is deformed by learning multi-scale visual and semantic relationships. The generated spatial guidance attention map $G$ and textual prompt $L_a$ are integrated into latent diffusion models with training-free guidance. The autoencoder-decoder in pixel space and the latent diffusion model in latent space retain original structure and parameters, and these components are frozen throughout the training process.}
	\label{fig:onecol}
\end{figure}

In this section, we present our proposed framework, denoted as ST-LDM, and the overall pipeline is depicted in \cref{fig:onecol}. 

In pixel space (\cref{section:sec3.1}), visual feature extraction and multi-modal localization are concurrently performed during the downsampling process, leveraging Swin Transformer~\cite{Liu_2021_ICCV} as the backbone. The feature extraction module acts as an autoencoder to concurrently reconstruct the image into a lower-dimensional latent space while furnishing hierarchical visual features for spatial localization. In parallel, the multi-modal Transformer module leverages the interplay between image and language to provide positional guidance for the denoising process. We innovatively introduce deformable feature alignment to dynamically adjust spatial constraints, ensuring that the generated object aligns with semantic context of the input image. The congruence in interaction between two branches is optimized through the application of a similar network structure (SwinT) to both of the aforementioned modules. 

In latent space (\cref{section:sec3.2}), appearance-based text prompt is separately mapped into denoising U-Net via cross attention. Subsequently, a region-wise backpropagation scheme is introduced to iteratively update the activations in the network to match the desired layout specified in pixel space. The fundamental structure and parameters of diffusion models remain constant, with trainable parameter layers exclusively concentrated in the pixel space. Consequently, the proposed ST-LDM emerges as a universal framework adaptable to a spectrum of pretrained LDMs.

\subsection{Hierarchical Feature Extraction}\label{section:sec3.1}
\subsubsection{Autoencoder with Swin-Transformer.} \label{section:sec3.1.1}
LDMs~\cite{nokey} leverages an individually pretrained autoencoding model for perceptual image compression, but its quantitative performance exhibits significant variability across different downsampling factors due to the inherent instability of its encoding structure. Recently, Swin-Transformer has emerged as a generic visual feature extractor, owing to its robust modeling capabilities~\cite{Liu_2021_ICCV}. It initially splits an input RGB image into non-overlapping patches with a 4$\times$4 size, and the resolution of subsequent layers' output continually decreases accordingly. However, excessive information compression will induce distortion of noisy images in the denoising process.

To address this issue, we add a trainable patch expanding layer to convert the selected feature map to the specified size. The original structure of Swin-Transformer is retained for hierarchical feature extraction, and an intermediate feature map $V_i$ is selected and upsampled to high-dimensional resolution via patch expanding layer. It can increase the resolution by a factor of 4 while simultaneously reducing the dimension to a configurable value ($H_i \times W_i\times C_i\rightarrow 4H_i \times 4W_i\times C_t$, $C_i\geq16C_t$). The layer consists of two operations, a linear layer to convert the feature dimension ($H_i \times W_i\times C_i\rightarrow H_i \times W_i\times 16C_t$), and a rearrange operation to expand the resolution to 4× and reduce the dimension to one-sixteenth ($H_i \times W_i\times 16C_t\rightarrow 4H_i \times 4W_i\times C_t$). 

The extracted feature map is optional within $\left\{V_2,V_3,V_4\right\}$ contingent upon the depth of compression perception. As the initial patch size in Swin-Transformer is 4$\times$4, the resulting output resolution for patch expansion remains fixed at 4$\times$ relative to its initial dimensions. Consequently, $\left\{V_2,V_3,V_4\right\}$ corresponds to downsampling factors $f=\left\{2,4,8\right\}$. Due to the varied channel configurations across model variants, the output channel dimension remains adaptable. The encoding-decoding process in pixel space exhibits a symmetrical structure, wherein the linear layer in patch expanding and patch partition undergoes low-cost training following~\cite{nokey} to cater to specific downstream tasks while retaining the formidable modeling capabilities of pretrained models.

\subsubsection{Multimodal Transformer.}\label{3.1.2}
We jointly embed language and vision features during visual encoding with Swin-Transformer to generate semantically coherent positional guidance for object generation. Consistent with previous works~\cite{Yang_2022_CVPR}, we densely integrate spatial-based language features $L_s\in R^{T_{L_s} \times C_{L_s}}$ into image features $M_i\in R^{H_i\times W_i\times C_i}$ via cross attention implementing
\begin{equation}
	\operatorname{Attention}(M^{q}_{i}, L^{k}_{s}, L^{v}_{s})=\operatorname{softmax}\left(\frac{M^{q}_{i}{L^k_{s}}^T}{\sqrt{d}}\right) \cdot L^{v}_{s}
\end{equation}
with $L^k_{s}= W^kL_s$, $L^{v}_{s}=W^vL_s$, $M^{q}_{i}=\operatorname{flatten}(W^{q}M_i)$, and $ L^k_{s}, L^{v}_{s}\in R^{T_{L_s} \times C_{i}}$, $M^{q}_{i}\in R^{H_iW_i \times C_i} $. $W^k$, $W^v$, $W^q$ are implemented as a 1$\times$1 convolution with $C_i$ number of output channels.

\subsubsection{Deformable Feature Alignment.} 
The TOG task presents a unique challenge in that the given image lacks the referred object, which is adverse to aforementioned attention mechanism. The input language encompasses both appearance and spatial attributes. If exclusively relying on spatial information, the model can only provide a generic localization of the region due to its limited grasp of object concepts, consequently compromising the semantic rationality of edited images (\eg a generated mouse being larger than the existing elephant). However, when appearance attributes are incorporated into the input, the cross-attention module faces a dilemma.  It is tasked with reconciling the existing spatial attributes with the absent appearance features in the image, leading to conflicting information.  This predicament significantly hampers the model's performance and generalization ability, especially in cases where the described appearance features are already present in the given image.

\begin{figure}[tb]
	\centering
	\includegraphics[width=1\linewidth]{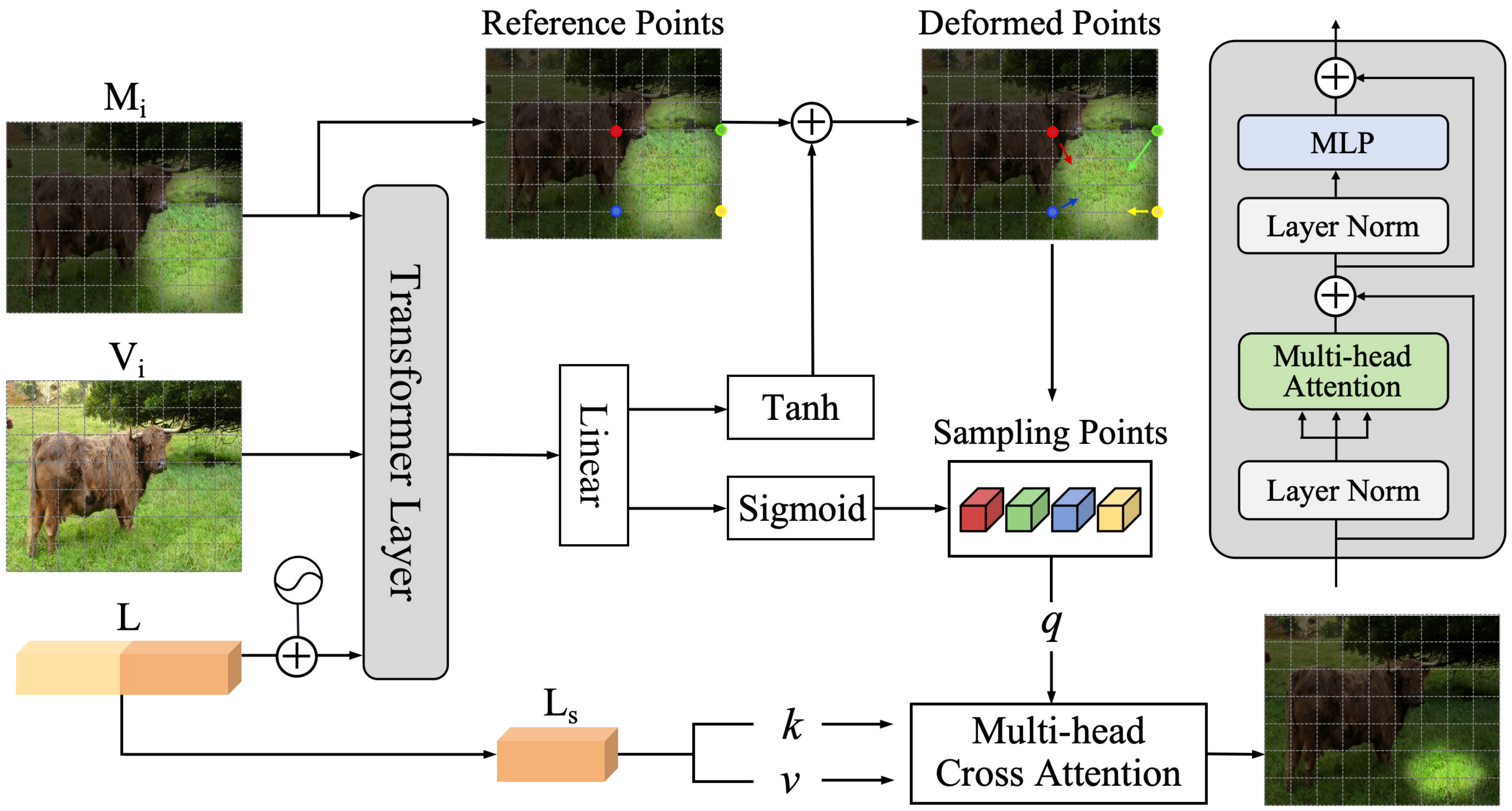}
	\caption{An illustration of deformable feature alignment. Global visual feature $V_i$, multimodal feature $M_i$, and text feature $L$ are concatenated and inputted into the Transformer layer to learn offsets and modulation scalar. The multi-head cross attention is then performed to generate the refined attention map with sampled queries in $M_i$ and language keys and values.}
	\label{fig:onecol1}
\end{figure}

Inspired by deformation mechanisms~\cite{8237351,Xia_2022_CVPR,zhu2021deformable}, we propose deformable feature alignment to mitigate potential conflicts by shifting candidate queries towards regions of interest in multi-head cross attention, as illustrated in \cref{fig:onecol1}. The multimodal map $M_i$ focuses on visually existing positional features by capturing the correlation between spatial-based textual and visual features, and it undergoes deformation with the alignment of object descriptions and multi-scale image features. 

\textbf{Offset Generation.} Initially, we add a sinusoidal 1D positional encoding $e_{pos}\in R^{T_L \times C_L}$ to the word embedding $L$, updating as $L\leftarrow L+e_{pos}$. A uniform grid of points $ p\in R^{H_i/\gamma \times W_i/\gamma\times 2}$ are generated as references on $M_i$, where $\gamma$ is the sampling factor. $M_i$, $V_i$, and $L$ are normalized to the same dimension through a linear layer and then concatenated as inputs into a transformer layer to generate patch-wise multimodal offset features. Subsequently, a linear projection is applied to modify the dimension to \small $H_i/\gamma \times W_i/\gamma\times 3$, \normalsize where the first two channels correspond to the learned 2D offsets $\Delta p$ constrained with maximum offset value $s_i$ ($\Delta p \leftarrow s_i \tanh (\Delta p)$), and the last channel is fed to a sigmoid layer to obtain modulation scalar $\Delta m$, which is introduced to exclude redundant context of which the elements lie in the range $[0,1]$.

\textbf{Multi-head Cross Attention.} 
For reference points $p$ on $M_i$, 2D offsets $\Delta p$ and modulation scalar $\Delta m$ are generated via offset generation module. $\Delta p$ refines the spatial position according to textual descriptions, and $\Delta m$ introduces an additional dimension of freedom to exclude irrelevant context. Both of them are learned from the same transformer layer and activated with distinct modules. We designate $\Delta p_k$ and $\Delta m_k$ as the offset and scalar for the $k$-th reference point $p_k$, where $k\in \{0,1,2,...,$ \small$ H_i W_i/\gamma ^2$\normalsize$\}$. As $p_k+\Delta p_k$ is fractional, a bilinear interpolation $\phi(I,p_k)$ is applied as 
\begin{equation}
	\phi\left(I,p_k\right)=\sum_{q}g\left(p_{kx}, q_{x}\right)\cdot g\left(p_{ky}, q_{y}\right)\cdot I(q),
\end{equation}
where $g(a, b)=\max (0,1-|a-b|)$ and $q$ indexes the 4 integral points closest to $p_k$. The sampling queries are deformed as 
\begin{equation}
	M^{q}_{i}=\operatorname{flatten}(W^{q}\phi (M_i,p+\Delta p))\label{8}.
\end{equation}
 The sampling visual queries are attended to the\label{key} language keys with modulate attention scores $\Delta m$. A matrix completion function $\psi(A)$ is then introduced to fill in the attention score of unsampled queries and restore the dimension of the attention map to \small$H_iW_i\times T_{L_s}$\normalsize. For each unsampled query $a_u$ in $A$, the attention score of $a_u$ is the weighted sum of scores of all the sampled queries $a_s$ in the range $\mathbb{R}$ around $a_u$. The number of $a_s$ in the range $\mathbb{R}$, denoted as $\operatorname{card}(a_s)$, reflects the degree of deformation near $a_u$ and is introduced as a modulation factor. $\operatorname{Avg}(\operatorname{card}(a_s))$ refers to the average number of $\operatorname{card}(a_s)$ for all the $a_u$ in $A$. 
\begin{equation}
	\psi(a_{u})=\frac{\sum_{a_{s}\in\mathbb{R}} w_sa_{s}}{\sum w_s}\cdot \frac{\operatorname{card}(a_s)}{\operatorname{Avg}(\operatorname{card}(a_s))}\label{9},
\end{equation}
\begin{equation}
	w_s=\frac{1}{\operatorname{d}(a_{u},a_{s})+\epsilon},
\end{equation}
where $\operatorname{d}(a_{u},a_{s})$ refers to the Euclidean distance of $a_{u}$ and $a_{s}$, and $\epsilon$ is a modulation
constant. The output of the $j$-th attention head is formulated as
\begin{equation}
	Z^{(j)}=\operatorname{softmax}\left(\psi(\frac{M^{q}_{i}{L^k_{s}}^T}{\sqrt{d}}\cdot \Delta m)\right) \cdot L^{v}_{s}.
\end{equation}

We preserve the original structure of cross attention in the initial fusion phase and utilize spatial conditions $L_s$ as the key and value to generate a coarse-grained attention region. In subsequent stages, we introduce deformable feature alignment with $s_i$ set as $\{8,4,2\}$ to progressively refine spatial positions during the process of deep feature extraction. This strategy can bias the attention to spatial position information rather than image local features, simultaneously with fast convergence and high computational efficiency. In the last layer, we directly extract the attention score in the multi-head cross attention and then convert it to a two-dimensional matrix averaged by the channel axis, denoted as $G$, as location guidance in denoising U-Net.
\subsection{Training-free Spatial Guidance}\label{section:sec3.2}
Location guidance $G$ encompasses spatial attention generated in pixel space, delineating the approximate boundary of the generation position. Despite the refinement conducted through deformable feature alignment which adaptively adjusts spatial localization based on multimodal information, the numerical difference of attention scores in $G$ may influence the shape of the generated object, compared to a manually defined bounding box. Thus, we utilize region-wise backward guidance to avoid potential overly aggressive intervention. 

LDMs~\cite{nokey} augment underlying UNet backbone with spatial transformer, where the core module is cross-attention with $L^{(\iota)}_{k}= W_kL_a$, $L^{(\iota)}_{v}=W_vL_a$, $V^{(\iota)}_{q}=\operatorname{flatten}(W_{q}z_t)$, where $\iota$ is the index of the relevant layer in the network, $L_a$ refers to appearance-based text prompt, and $z_t$ is the latent representation of the input image. The attention weight matrix is described as:
\begin{equation}
	S^{(\iota)}=\operatorname{softmax}\left(\frac{V_{q}^{(\iota)}{L^{(\iota)}_{k}}^T}{\sqrt{d}}\right). 
\end{equation}

We extract the first cross-attention block in the up-sampling branch to compute backpropagation, which yields superior performance while incurring minimal computational time overhead~\cite{chen2023trainingfree}. The corresponding attention score at timestamp $t$ is denoted as $S_t$, and it has a similar compression scale with $G$, which is advantageous in subsequent calculations pertaining to the energy function. The energy function is defined as
\begin{equation}
	E(S_t,G) =\left(1-\frac{\sum(\theta(\hat{G},\beta)\odot S_t)}{\sum{S_t}}\right)^2\label{con:inventoryflow},
\end{equation}
where $G$ is converted to the same resolution as $S_t$, denoted as $\hat{G}$. $\theta(\hat{G},\beta)$ is an activation function with two steps, defined as
\begin{equation}
	\theta(\hat{G},\beta)=\operatorname{Dilate}(\hat{G})\leftarrow\hat{G}_{mn}  =\begin{cases}
		1& \text{$\hat{G}_{mn}\ \textgreater \ \beta$ }\\
		0& \text{$\hat{G}_{mn}\leq \beta$ }
	\end{cases},
\end{equation}
where the element $\hat{G}_{mn}$ is first activated with the threshold value $\beta$, and then $\hat{G}$ is dilated to contour external maxmatrix to eliminate the numerical interference on the shape of the object. At each denoising U-Net, the latent $\boldsymbol{z}_t$ is updated with the gradient of Eq. \ref{con:inventoryflow}.

\begin{equation}
	\boldsymbol{z}_t \leftarrow \boldsymbol{z}_t-\eta\sqrt{\left(1-\alpha_t\right) /  \alpha_t}  \nabla_{\boldsymbol{z}_t} E(S_t,G),
\end{equation}
where $\eta\textgreater0$  defines the strength of guidance and $\alpha_t$ refers to the decreasing sequence of noise schedule in stable diffusion.

\section{Experiments}

\subsection{Dataset}
Our primary focus of chosen datasets is the linguistic diversity of spatial conditions rather than text2img prompts since our framework is explicitly tailored for spatial guidance with the parameter layers of LDMs held frozen.

\textbf{VrR-VG.} VrR-VG~\cite{liang2019vrr} is constructed from Visual Genome~\cite{krishna2017visual} and contains 58,983 real images with 23,375 relation pairs. Excluding positional and statistically biased relationships, It encompasses more visually relevant relationships, like ``\textit{sit on}'', ``\textit{other side of}'', ``\textit{fly in}'', \etc. We select spatial relationships and restructure linguistic expressions to align with our input format. For instance, for annotations involving attributes (\textit{fire hydrant}, \textit{yellow}), attributes (\textit{man}, \textit{standing}), and beside (\textit{man}, \textit{fire hydrant}), we reformulate them as ``\textit{standing man} [\textit{beside yellow fire hydrant}]''. Subsequently, We remove the specific object and complete images employing image inpainting techniques~\cite{dong2022incremental} as the input image, and the ground truth is the original position of the removed object.

\textbf{ComCOCO.} ComCOCO~\cite{unpubl} is derived from RefCOCO+~\cite{kazemzadeh2014referitgame} and consists of 136,495 referring expressions for 34,615 objects in 23,951 image pairs. It is a synthetic dataset and the spatial relationship is delineated in the process of manual inspection. The annotations within ComCOCO encompass both appearance and spatial attributes, and we combine them in a manner analogous to the approach detailed above. We utilize real images in ComCOCO as the input image, and the corresponding synthetic images provide the ground-truth position.

\subsection{Implementation Details}
We utilize SwinV2-T~\cite{liu2022swin} with layer number $= \{2,2,6,2\}$ pretained on ADE20K~\cite{zhou2019semantic} as the autoencoder in pixel space. Text embedding is initialized with official weights of ALBERT Base v2~\cite{DBLP} with 11M parameters and 12 layers. Our framework is adaptable to various latent diffusion models, and we conduct experiments by building upon a pretrained Stable-Diffusion (SD) V-1.5~\cite{nokey} trained on the LAION-5B dataset~\cite{schuhmann2022laion5b}, unless stated otherwise. 

The training process is separated into two distinct phases. First, we train patch expanding and patch partition module on OpenImages~\cite{kuznetsova2020open} with KL-regularization following \cite{nokey} to ensure optimal scaling properties of the pretrained autoencoder concerning spatial dimensionality when applied to LDMs. Subsequently, we train multimodal fusion transformer layers with the backbone of Swin-T on a combined dataset comprising a modified version of VrR-VG and ComCOCO. VrR-VG contributes a diverse and balanced set of spatial relationships, albeit with potential inpainting artifacts. To mitigate this limitation, we introduce ComCOCO which employs real images as its input source. Moreover, 
spatial descriptions in ComCOCO predominantly revolve around the holistic perspective of an image, whereas VrR-VG is specifically directed toward capturing and describing the intricate relationships between objects within the visual content.

All the experiments are implemented on 4 V100 GPUs of 32G RAM. We set the downsampling factor $f$ = 4 and channel dimension $c$ = 3 in the autoencoder. Backward guidance is performed during the initial 10 steps of the diffusion process and repeated 3 times at each step with the strength factor $\eta$ set to 35. The input image is scaled to 480$\times$480, and the data augmentation strategy includes random cropping and flipping. The optimizer is AdamW~\cite{loshchilov2017decoupled} with $\beta_1=0.85$, $\beta_2=0.91$. The weight decay factor and the learning rate of the initial exponential decay are set to 0.003 and 0.00024, respectively, and the remaining parameters are randomly initialized. The model is trained for 172 rounds with the batch size set to 32 to achieve optimal performance.

\subsection{Qualitative Evaluation}
\begin{figure}[htbp]
	\centering
	\includegraphics[width=1\linewidth]{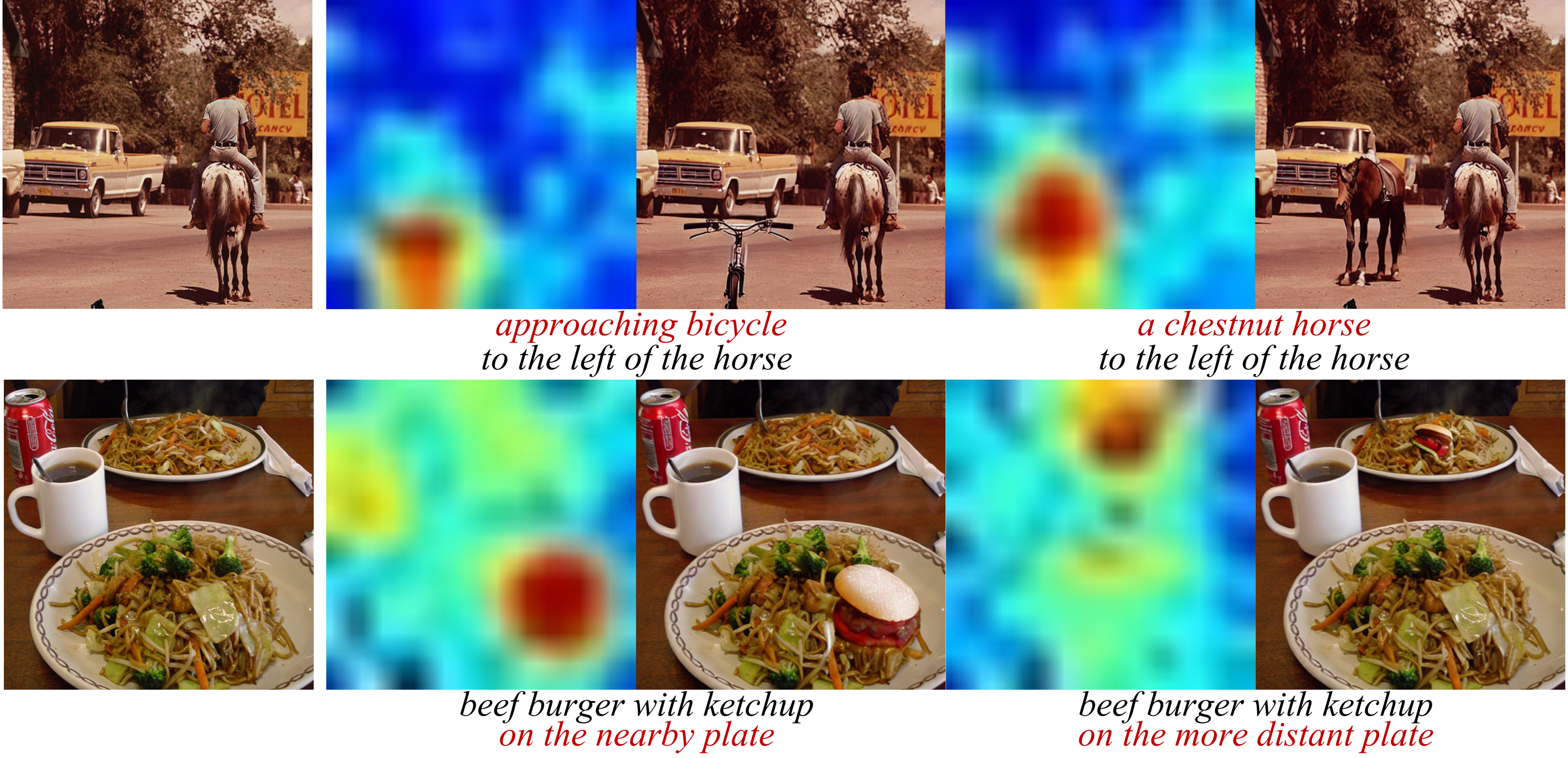}
	\caption{The first column corresponds to the input real image, and we visualize the guidance attention map $G$ and corresponding edited results under different prompts with identical conditions (top) and identical prompts with different conditions (bottom).}
	\label{fig:onecol4}
\end{figure}

\begin{figure}[htbp]
	\centering
	\includegraphics[width=1\linewidth]{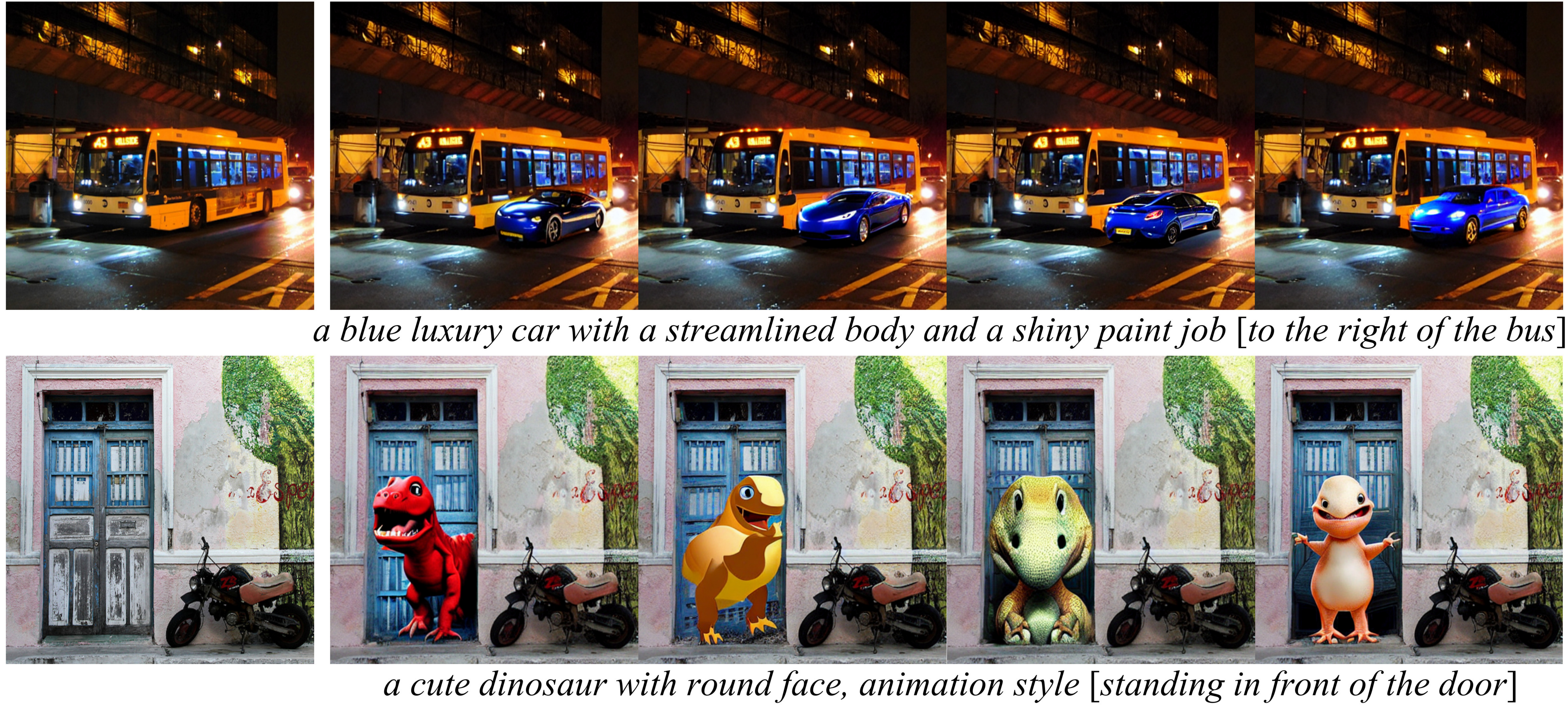}
	\caption{Various results produced by ST-LDM, showcasing diverse appearances while adhering to the provided text descriptions.}
	\label{fig:onecol5}
\end{figure}

We apply our method on a multitude of real images from various domains, as exhibited in \cref{fig:onecol3}. We further compare the localization ability in \cref{fig:onecol4} and generation ability in \cref{fig:onecol5}. \Cref{fig:onecol4} shows the qualitative evaluation of ablative comparison of appearance and spatial properties. The deformable feature alignment module exhibits a remarkable ability to adjust spatial positions in accordance with linguistic descriptions and visual context. This dynamic adjustment imparts a direct influence on the image editing process within the latent diffusion model. As illustrated in \cref{fig:onecol5}, our method maintains the robust generative capabilities intrinsic to diffusion models with the preservation of semantics coherence to the original scene. The spatial constraints exerted on the denoising process have no discernible impact on the diversity and fidelity of the generated objects owing to region-wise backward guidance. More visualizations are provided in \cref{Appendix C}.

	\vspace{-0.56cm} 
\subsection{Quantitative Comparison}
\subsubsection{Evaluation Metrics.}\label{metric}We conduct quantitative assessments of the quality of edited images by examining two facets. (1) Object alignment: We introduce VISOR~\cite{gokhale2022benchmarking} to score spatial object relationships according to text descriptions and text2img metric CLIPScore~\cite{hessel2022clipscore} to measure the consistency of generated objects with the input text. (2) Image fidelity: We utilize full-reference image quality assessments FID~\cite{zhang2018unreasonable} and LPIPS~\cite{Zhang_2018_CVPR} to measure image fidelity. HPS~\cite{wu2023human} is also introduced to align with human preference.

\subsubsection{Benchmark Dataset.} We construct a benchmark dataset for quantitative analysis and compare ST-LDM with related tasks. We find a distinct bias in the textual descriptions associated with the presence or absence of the specific object within the image. When the object is present in an image, the descriptions tend to emphasize distinctive features that differentiate the object from other visual entities. Conversely, when generating an object that does not exist in the original image, the focus shifts towards the detailed attributes of the object. Therefore, we randomly select 1,500 images from COCO2014~\cite{lin2014microsoft}, and manually describe the object required to be generated in each image, simultaneously grounding the ideal position using bounding boxes. The detailed construction process is illustrated in \cref{B}.
\subsubsection{Results.} We compare our method with editing models conditioned by diverse modalities in \cref{tab:example1}. Existing text-guided editing models exhibit limitations in generating new objects within complex scenes, and our model outperforms by a significant margin benefiting from explicit guidance generated from the deformable feature alignment. When compared to approaches employing forced modalities to precisely confine generated objects to designated regions, our method exhibits the capacity to edit local region of real images under the weak supervision of language, and the generated objects accurately adhere to the spatial relationships as specified in the text conditions.

\begin{table}[h]
	\vspace{-0.3cm}
	\caption{Comparison with editing models conditioned by diverse modalities.}
	\setlength{\tabcolsep}{2.8pt}
	\label{tab:example1}
	\centering
	\begin{tabular}{ccccccccccccc}
		\toprule
		{Condition}&{Model}  &FID$\downarrow$&LPIPS$\downarrow$&HPS$\uparrow$&CLIPScore$\uparrow$&VISOR$\uparrow$\\  
		\midrule
		Text&InstructPix2Pix~\cite{brooks2023instructpix2pix}&22.25&30.93&41.58&60.45&72.15\\
		Text&SmartEdit~\cite{huang2023smartedit}&18.47&27.11&47.40&69.02&77.86\\
		Mask&BlendedLatentDiffusion~\cite{Avrahami_2023}&12.93&18.05&54.33&80.96&88.12\\
		Mask&PaintByWord~\cite{andonian2023paint}&9.72&20.68&50.26&81.23&87.41\\
		BBox&BreakAScene~\cite{avrahami2023break}&7.89&16.24&59.25&81.10&91.95\\
		BBox&GLIGEN~\cite{li2023gligen}&6.24&17.09&62.41&80.62&90.53\\
		\midrule
		Text& ST-LDM &\textbf{6.16}&\textbf{15.54}&\textbf{64.28}&\textbf{81.97}&\textbf{92.39}\\
		\bottomrule
	\end{tabular}
	\vspace{-0.3cm}
\end{table}

\subsection{Ablation Study}\label{4.5}
\begin{figure}[tb]
	\centering
	\includegraphics[width=1\linewidth]{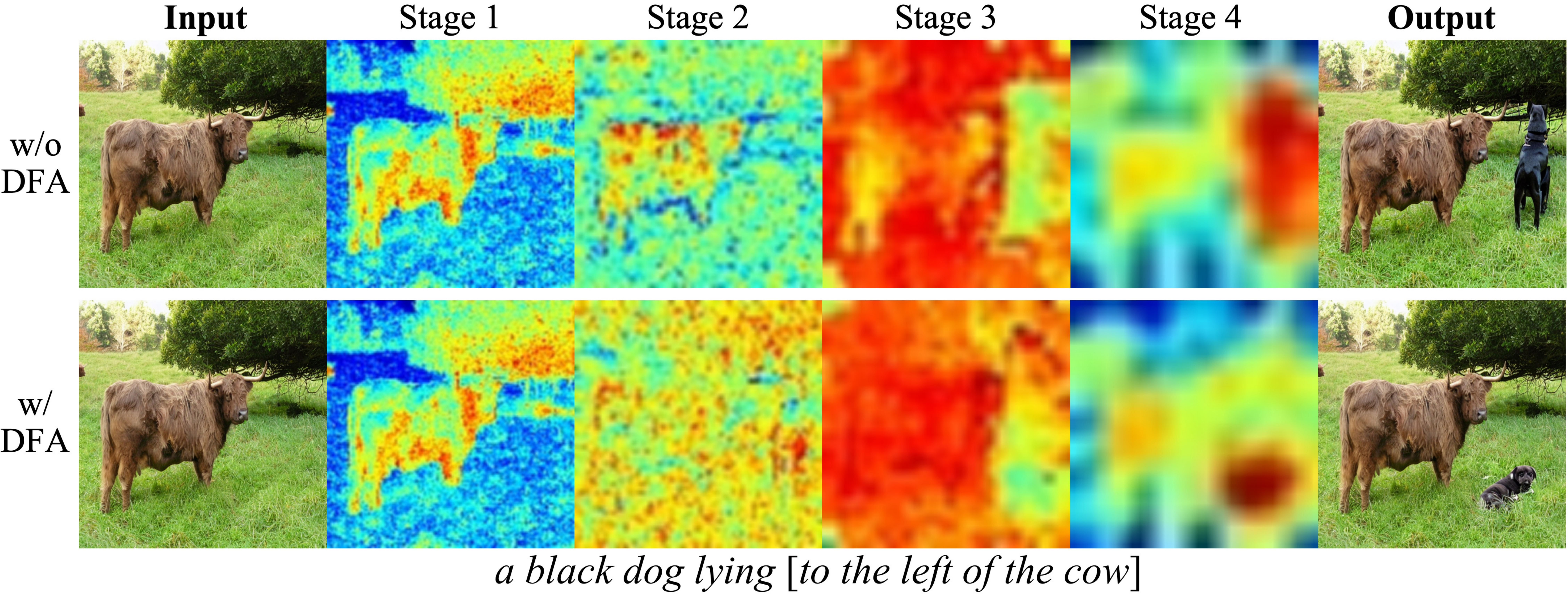}
	\caption{Ablative visualization of attention maps at diverse stages with DFA.}
	\label{fig:onecol6}
	\vspace{-0.1cm}
\end{figure}

\textbf{Deformable Feature Alignment.} We replace the feature alignment of multimodal fusion Transformer with our deformable mechanism at different stages, resulting in a direct impact on the modification of generated regions, as depicted in \cref{fig:onecol6}. We introduce FID and LPIPS to assess the extent of alteration in the original image. A user study is also conducted to assess the appropriateness of object position, which serves as a more persuasive and reference-worthy metric. As delineated in \cref{tab:example5}, the integration of deformable feature alignment refines the attention region, resulting in a numerical improvement across all metrics. However, premature intervention of DFA in the initial stage interferes with spatial perception (since the input $M_1$ of DFA will be equal to $V_1$ if we introduce DFA in Stage1). Actually, in the initial stage of modal fusion, the model generates a preliminary spatial positioning via cross attention, and DFA is designed as a refining strategy. More illustrations are provided in \cref{A}.

\begin{table}[h]
\caption{Ablation study of deformable feature alignment (DFA) applied at different stages. Users are required to estimate the conformance of the object's position (True or False), and the score is calculated based on the overall accuracy across all the images. IT refers to the inference time (sec/image).}
\label{tab:example5}
	\centering
		\setlength{\tabcolsep}{7pt}
	\begin{tabular}{cccc|ccccc}
		\toprule
		\multicolumn{4}{c|}{Stages w/ DFA}	&\multirow{2}{*}{UserScore$\uparrow$} &\multirow{2}{*}{FID$\downarrow$}&\multirow{2}{*}{LPIPS$\downarrow$}&\multirow{2}{*}{IT}\\ 
		S-1&S-2&S-3&S-4    \\
		\hline
		\checkmark&\checkmark&\checkmark&\checkmark&89.3&7.50&16.23&$\sim$7.0\\
		&\checkmark&\checkmark&\checkmark&\textbf{92.4}&\textbf{6.16}&\textbf{15.54}&$\sim$6.5\\
		&&\checkmark&\checkmark&91.1&9.14&20.46&$\sim$6.3\\
		&&&\checkmark&82.5&15.97&25.43&$\sim$6.1\\
		&&&&75.9&27.03&34.86&$\sim$6.0\\
		\bottomrule
	\end{tabular}
	
\end{table}

\textbf{Backward guidance.} We further assess the effectiveness of backward guidance in \cref{tab:example6}. We replace guidance maps with textual conditions during the denoising process, and the notable decline indicates that linguistic information inadequately constrains the spatial placement of generated objects. We employ an activation function to eliminate potential discrepancies in attention scores, followed by dilation to mitigate the morphological interference of the attention region to the object.

\begin{table}[h]
	\caption{Ablation study of backward guidance. Users are required to assess the matching score between the edited image and spatial descriptions. IT refers to the inference time (sec/image).}
	\label{tab:example6}
		\setlength{\tabcolsep}{6pt}
	\centering
	\begin{tabular}{ccccccccccccc}
		\toprule
		&\multirow{2}{*}{UserScore}&\multirow{2}{*}{CLIPScore}& \multicolumn{2}{c}{VISOR}& \multirow{2}{*}{IT}	\\  \cmidrule(r){4-5}  
		&&&uncond    &cond  \\
		\midrule
		w/o guidance&58.73&55.07&11.54&60.45&$\sim$4.1\\
		w/o activation&80.86&69.01&37.26&90.94&$\sim$6.0\\
		w/o dilation&83.98&70.52&39.14&90.37&$\sim$6.2\\
		\midrule
		Ours&\textbf{89.11} &\textbf{81.97}&\textbf{43.41}&\textbf{92.39}&$\sim$6.5\\
		\bottomrule
	\end{tabular}
	\vspace{-0.5cm}
\end{table}
\section{Conclusion}
In this paper, we explore the text-grounded object generation in real-world images. A universal framework with deformable feature alignment is proposed to hierarchically refine the attention region conforming to vision and language features. The generated guidance map can be incorporated into any latent diffusion model with training-free backward guidance. Our model addresses the inherent limitation of diffusion models in spatial perception under textual supervision, and offers valuable insights for subsequent research.
\clearpage
\bibliographystyle{splncs04}
\bibliography{main}
\clearpage
\begin{appendix}
\title{ST-LDM: A Universal Framework for Text-Grounded Object Generation in Real Images\\ Supplementary Material}
	\section{Deformable Feature Alignment}\label{A}
	In this section, we make an exhaustive exposition of deformable feature alignment. Most previous works for referring image segmentation employ cross attention mechanisms to model the relationship between language and image. This approach highlights the visual entity pertinent to textual descriptions, wherein the incorporation of deformable mechanisms serves to accommodate geometric variations of objects and direct attention towards salient regions enriched with relevant features. In the TOG task, the region of interest is devoid of specific visual entities, instead emphasizing conformity with linguistic cues and compatibility with other visual entities. Consequently, our proposed deformable strategy differs fundamentally from previous deformable mechanisms.
	
\begin{figure}[h]
		\centering
		\includegraphics[width=1\linewidth]{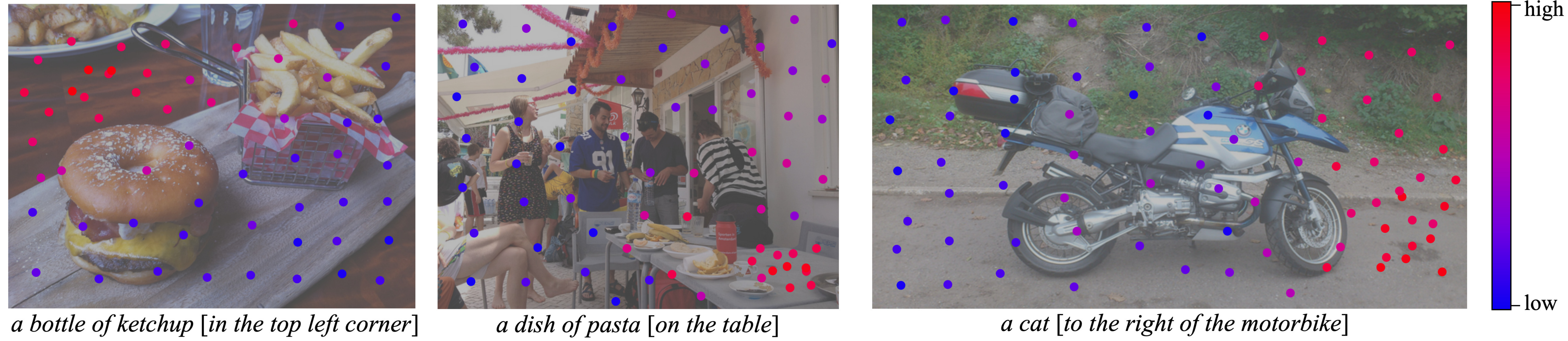}
		\caption{Visualization of deformed queries. We draw the deformed points and modulation scalar from multi-head feature maps at the last stage. Each sampling point is marked as a filled circle whose color indicates its corresponding modulation scalar.}
		\label{fig:onecol11}
		\vspace{-0.5cm}
	\end{figure}

	We further visualize the deformed queries with modulation scalars in \cref{fig:onecol11}. The color distribution and spatial density of small circles reflect the differentiation in spatial-based and appearance-based language, aligning with the influence exerted by modulation scalars and offsets.
	
	\vspace{-0.4cm}
	\subsection{Detailed Architectures}
	We utilize Swin-Transformer as the backbone and introduce deformable feature alignment at the last three layers. The detailed architectures specifications are reported in \cref{tab:example7}.
	
	\vspace{-0.2cm}
		\begin{table}[h]
		\caption{Architecture specifications of deformable multimodal Transformer. $\boldsymbol{N}$: layer number. $\boldsymbol{C}$: channel dimension. \textbf{win. sz.}: window size. \textbf{heads}: number of multi heads. }
		\label{tab:example7}
		\renewcommand{\arraystretch}{1}
		\setlength{\tabcolsep}{3pt}
		\centering
		\begin{tabular}{c|c|c|c}
			\toprule
			Stage 1&Stage 2&Stage 3 &Stage 4\\  
			(56$\times$56)&(28$\times$28)&(14$\times$14)&(7$\times$7)\\
			\hline
			$N$=2, $C$=96&$N$=2, $C$=192&$N$=6, $C$=384&$N$=2, $C$=768\\
			win. sz. 7$\times$7&win. sz. 7$\times$7&win. sz. 7$\times$7&win. sz. 7$\times$7\\
			heads: 3&heads: 6&heads: 12&heads: 24\\
			&$\gamma$=4, $s_i$=8&$\gamma$=2, $s_i$=4&$\gamma$=1, $s_i$=2\\
			&$\mathbb{R}$: 8$\times$8&$\mathbb{R}$: 4$\times$4&$\mathbb{R}$: 2$\times$2\\
			\bottomrule
		\end{tabular}
	\end{table}
	\subsection{Computational Resources}
Parameters and computational resources of ST-LDM compared with our selected backbone are shown in \cref{tab:example199}.

	\begin{table}[h]
		\caption{Parameters and Computational resources of ST-LDM. }
		\label{tab:example199}
		\centering
		\setlength{\tabcolsep}{8pt}
		\begin{tabular}{ccccc}
			\toprule
			&\# param. & FLOPs &MACs&Inference Time\\
			\midrule
			Swin-T & 28M & 4.5 G&6.3G&-\\
			SD V-1.5 & 859M &1587G&2731G &$\sim$ 4.1 s/img\\
			\midrule
			Ours &941M&1660G&2985G&$\sim$ 6.5 s/img \\
			\bottomrule
		\end{tabular}
		\vspace{-0.5cm}
	\end{table}	
	\subsection{Ablative Analysis}
	We assess the effectiveness of deformable feature alignment in \cref{4.5} and further analyze its internal modules in this section. DFA is essentially a two-stage attention mechanism. We first concatenate three features of disparate scales and utilize self-attention in Transformer layer to generate sampling queries with modulation scalar. Subsequently, cross attention is conducted on the sampled visual features and spatial language to refine the attention region. 
	
	The ablation study of essential modules within DFA is presented in \cref{tab:example8}. While the benchmark dataset provides manually labeled ground truth, the TOG task is a typical ill-posed problem and traditional metrics (\eg IoU) can not comprehensively measure the performance of the model. To mitigate this limitation, we introduce the discrepancy of size and center point between the generated object and the ground truth. The discrepancy is mapped to a numerical score ranging from 0 to 100, wherein a smaller discrepancy corresponds to a higher score.
	\begin{table}[h]
	\caption{Ablation study of DFA modules based on our benchmark dataset. \textbf{Attn}: T and C refer to the transformer layer and multi-head cross attention, respectively. T-only signifies that the attention map is generated directly from the Transformer layer, and C-only excludes DFT and retains the backbone structure. \textbf{T$_{\operatorname{input}}$}: the input modality of transformer layer. \textbf{ofs.}: 2D offsets $\Delta p$. \textbf{sca.}: modulation scalar $\Delta m$. \textbf{card.}: the modulation factor in \cref{9}. \textbf{Size}: size discrepancy between the generated object and GT. \textbf{Dist.}: distance between center points of the generated object and GT.}
	\label{tab:example8}
	\renewcommand{\arraystretch}{1.1}
	\setlength{\tabcolsep}{6pt}
	\centering
	\begin{tabular}{ccccc|cccc}
		\toprule
		Attn.&T$_{\operatorname{input}}$&ofs.&sca.&card.&VISOR&Size&Dist.&IoU\\  
		\hline
		{T+C}&$M_i$+$L$&\checkmark &\checkmark&\checkmark&77.55&46.25&68.78&34.01 \\
		{T+C}&$V_i$+$L$&\checkmark&\checkmark&\checkmark&79.43&51.46&63.92&38.67\\
		{T+C}&$V_i$+$M_i$&\checkmark&\checkmark&\checkmark&76.23&40.08&62.57&29.19\\
		\hline
		{T+C}&\checkmark&\ding{55}&\checkmark&\checkmark&87.23&68.02&73.14&49.88\\
		{T+C}&\checkmark&\checkmark&\ding{55}&\checkmark&83.54&76.30&71.61&52.49\\
		{T+C}&\checkmark&\checkmark&\checkmark&\ding{55}&89.27&83.36&76.43&61.07\\
		\hline
		T&\checkmark&\ding{55}&\ding{55}&\ding{55}&67.54&34.36&52.81&28.95\\
		C&\ding{55}&\ding{55}&\ding{55}&\ding{55}&74.99&44.13&68.37&32.86\\
		\hline
		{T+C}&\checkmark&\checkmark&\checkmark&\checkmark&\textbf{92.39}&\textbf{89.42}&\textbf{77.90}&\textbf{65.12}\\
		\bottomrule
	\end{tabular}
	
\end{table}
	\section{Construction of Benchmark Dataset}\label{B}
	The ideal benchmark adheres to two stringent criteria: firstly, the input images must be complete and real-world, thereby obviating the potential influence of artifacts; secondly, the input text must describe the object to be generated rather than already existing in the image. Given the absence of existing datasets that conform to these specific criteria, we undertake the meticulous manual construction of the benchmark dataset.
	
\begin{figure}[h]
		\centering
		\includegraphics[width=1\linewidth]{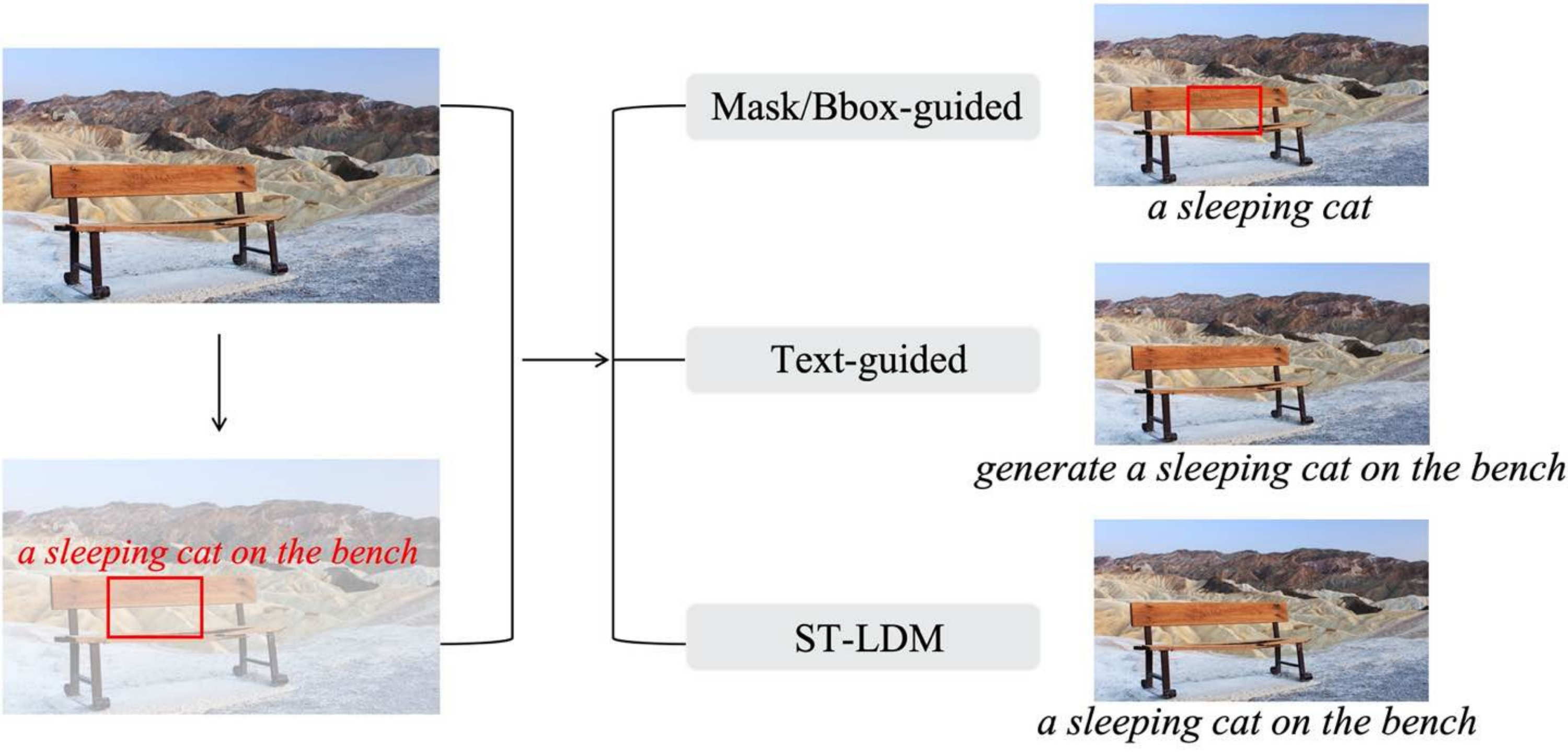}
		\caption{The illustration of construction process of an image pair and evaluating methods for distinct models.}
		\label{fig:onecol7}
	\end{figure}

	As depicted in \cref{fig:onecol7}, we randomly select an image in COCO2014 and manually annotate a sentence delineating the appearance and spatial attributes of the generated object, coupled with a visually reasonable reference region represented by bounding boxes. ST-LDM is the pioneering approach tailored for the novel TOG task and thus compared with relevant tasks under similar conditions. Specifically, we input the complete sentence into T2I generation models and measure the object alignment with language. For editing models conditioned by diverse modalities, we input the appearance prompt for the generated object followed by the annotated bounding box as conditional modality to evaluate the capacity of image fidelity.

\begin{figure}[h]
		\centering
		\includegraphics[width=1\linewidth]{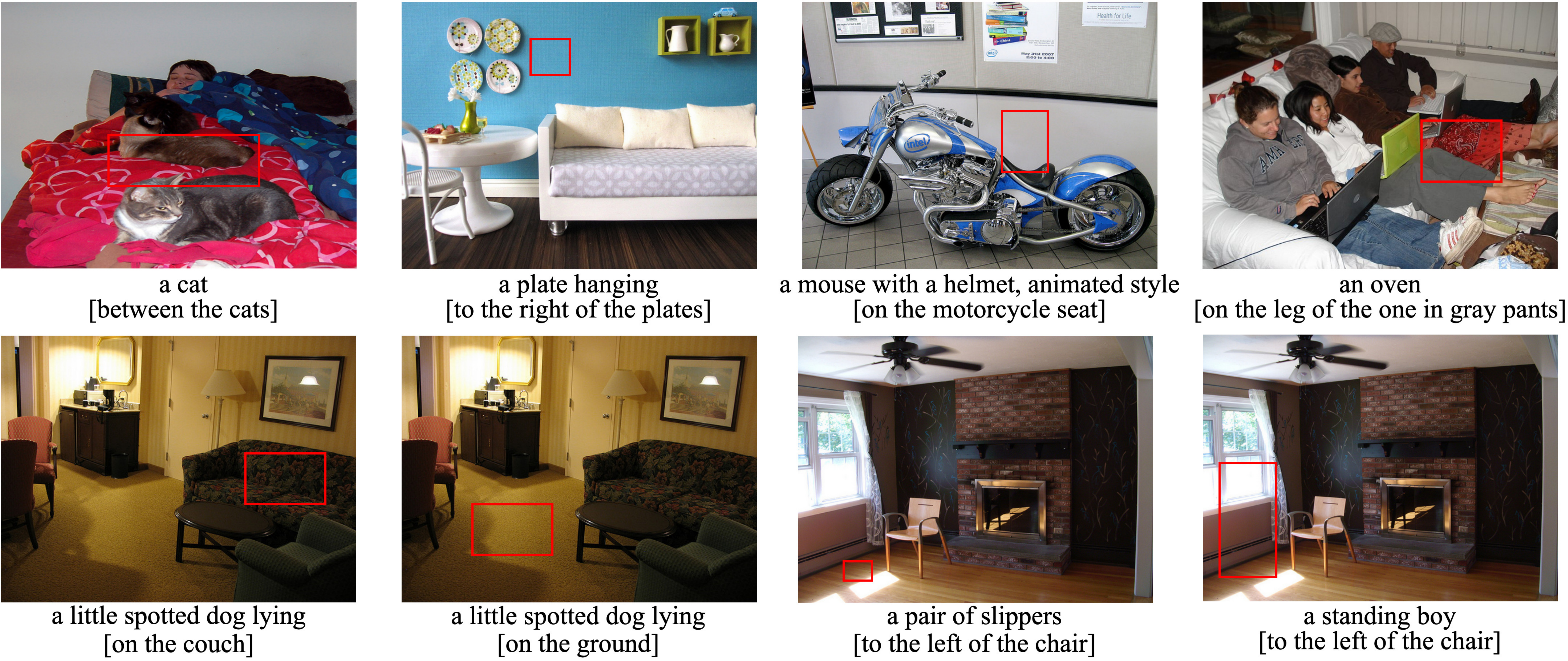}
		\caption{Exhibition of partial images in the benchmark dataset.}
		\label{fig:onecol8}
	\end{figure}

	We exhibit partial images of the benchmark in \cref{fig:onecol8}. To enhance the complexity of scene comprehension, the generated object is deliberately designed to possess comparable properties to other entities in the original image, and it may not accord with visual relationships in the real world (first row). Concurrently, we control the invariance of appearance and spatial descriptions respectively to assess the model's proficiency in learning language information from different dimensions (second row). The benchmark dataset comprises 1,500 images with 2,733 sentences and will be made publicly accessible soon for further research.

	\section{More Visualizations}\label{Appendix C}
\begin{figure}[h]
		\centering
		\includegraphics[width=1\linewidth]{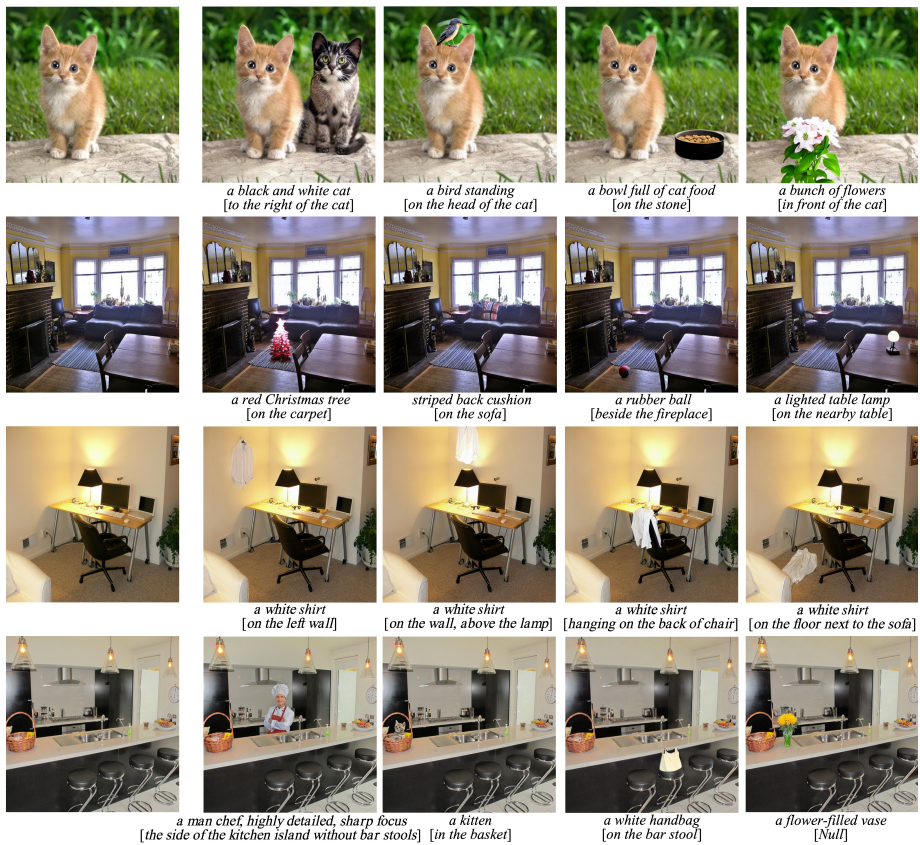}
		\caption{Editing results for an input image with diverse textual descriptions.}
		\label{fig:onecol9}
	\end{figure}

\begin{figure}[t]
		\centering
		\includegraphics[width=1\linewidth]{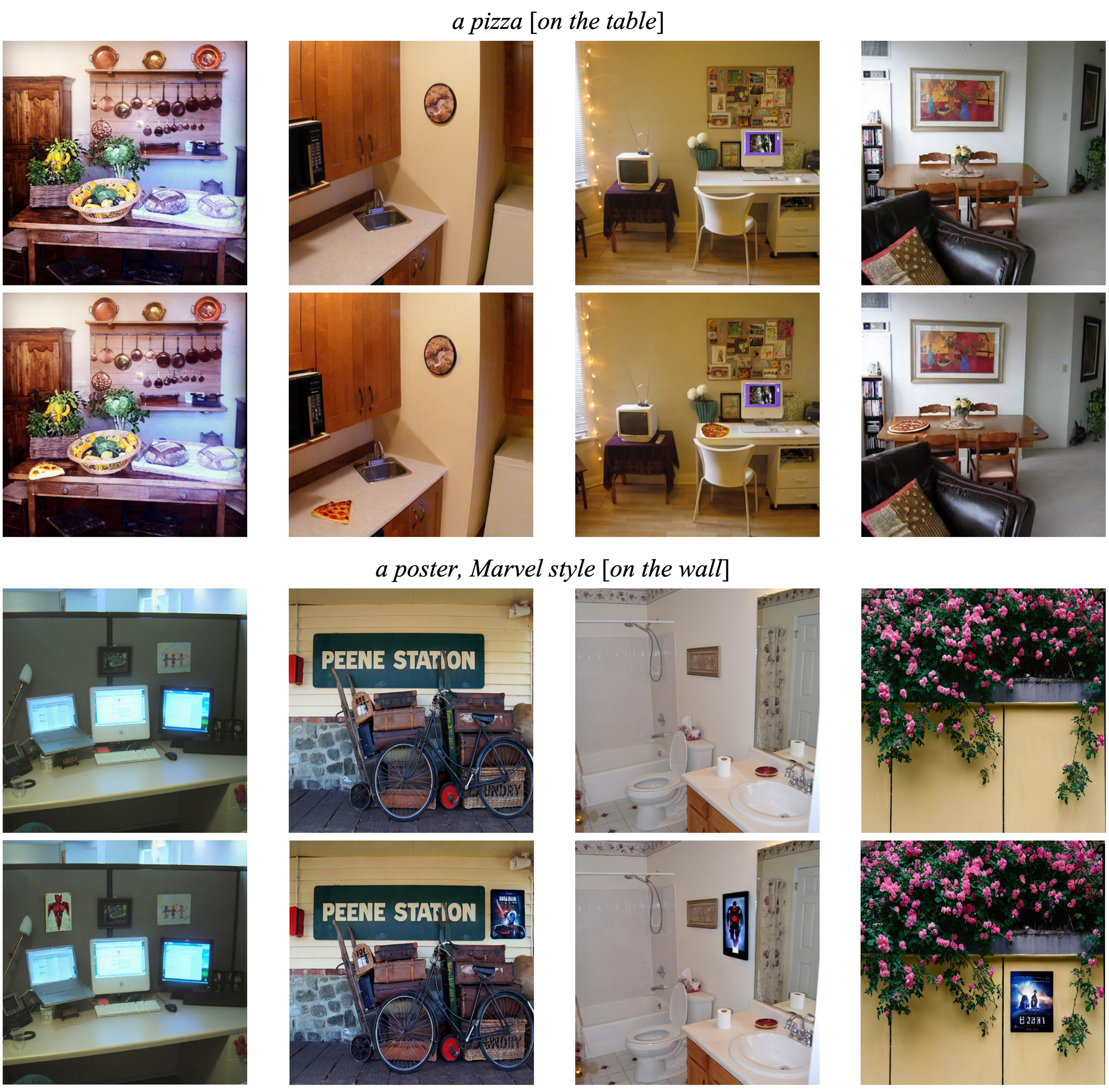}
		\caption{Editing results for a descriptive sentence with diverse input images.}
		\label{fig:onecol10}
	\end{figure}

	More qualitative results are shown in \cref{fig:onecol9,fig:onecol10}. We find an intriguing phenomenon that the model tends to generate objects within regions devoid of foreground features, which aligns with thinking patterns of human being. As exemplified in \cref{fig:onecol10}, we tend to place a pizza (poster) on the empty space of the table (wall). This characteristic is attributed to the structural design and training dataset of our model.
	
	\section{Limitations and Future Work}
	\begin{itemize}
		\item Definition of TOG task. Different from existing editing scenarios that mainly focus on modifying visual attributes of existing entities, the TOG task is specifically formulated to introduce entirely new visual entities into real-world images. There exists an overlap between the two tasks when the generated object affects the appearance of original entities. For example, if we want to generate a hat on the pre-existing man, it can be defined either as modifying visual attributes of the man or as generating a new object in the specific place. This paper will not discuss the above cases since we emphasize offering a valuable supplement to image editing scenarios. 
		\item Specific forms of statements. Our method requires specific forms of statements (appearance prompt and spatial conditions) as the seperated input. Separating two forms is crucial for subsequent modules, which is shown in \cref{fig:onecol}. As for integrated statements, we consider leveraging the preceding NLP module to extract the two forms of information. 
		\item Slight change of irrelevant region. In experiments, we find that there exist discernible disparities tightly in the vicinity of the generated object compared to the original pixels, which arises from the reconstruction of surrounding contextual pixel information during the denoising process of diffusion models. This can be ameliorated by post-processing techniques such as segmenting the generated object and then superimposing the original background information. This paper excludes an extensive discussion of this problem, as our primary objective is to explore a universal solution for the TOG task without bells and whistles. Our framework fundamentally follows a two-stage approach that converts textual supervision into spatial constraint, and in future work, we will explore the method to facilitate the direct incorporation of textual conditions into the denoising process of real images while preserving the fidelity of irrelevant pixels.
	\end{itemize}

\end{appendix}
\end{document}